\documentclass[letterpaper, 10 pt, conference]{ieeeconf}  
\IEEEoverridecommandlockouts                              
\usepackage{graphicx}
\usepackage{amsmath} 
\usepackage{fancyhdr,graphicx,amsmath,amssymb}
\usepackage[linesnumbered,ruled,vlined]{algorithm2e}
\usepackage{bm}
\usepackage{mathtools}
\usepackage{booktabs}
\usepackage{tabularx}
\usepackage{multirow}
\usepackage[subrefformat=parens]{subcaption}
\usepackage{color}
\usepackage{balance}
\usepackage{textcomp}
\usepackage{cite}
\newcommand{\EQREF}{Eq.~\eqref}
\newcommand{\FIGREF}{Fig.~\ref}
\def\proposed{L2B-SARL} 
\def\fixcolor{black}
\def\hstate{\bm {\tilde s}}
\def\rstate{\bm s}
\def\jstate{\bm s^{jn}}
\def\jpolicy{\overrightarrow{\pi}}
\def\vpref{v_{\text {pref}}}

\def\sup#1{^{(\rm #1)}}
\def\sub#1{_{\rm #1}}
\def\supi#1{^{(#1)}}
\def\vct#1{\mbox{\boldmath $#1$}}
\def\eg{{\it e.g.}}

\def\ie{{\it i.e.}}

\newcommand{\argmax}{\mathop{\rm argmax}\limits}

\title{\LARGE \bf
L2B: Learning to Balance the Safety-Efficiency Trade-off in \\Interactive Crowd-aware Robot Navigation
}

\author{Mai Nishimura$^{1}$ and Ryo Yonetani$^{1}$
\thanks{$^{1}$ Mai Nishimura and Ryo Yonetani are with OMRON SINIC X Corporation, Hongo 5-24-5, Bunkyo-ku, Tokyo, Japan.\newline 
        {\tt\small \{mai.nishimura,ryo.yonetani\}@sinicx.com}}
}

\begin{document}

\maketitle
\thispagestyle{empty}
\pagestyle{empty}

\begin{abstract}
This work presents a deep reinforcement learning framework for interactive navigation in a crowded place. Our proposed \emph{Learning to Balance (L2B)} framework enables mobile robot agents to steer safely towards their destinations by avoiding collisions with a crowd, while actively clearing a path by asking nearby pedestrians to make room, if necessary, to keep their travel efficient. We observe that the safety and efficiency requirements in crowd-aware navigation have a trade-off in the presence of social dilemmas between the agent and the crowd. On the one hand, intervening in pedestrian paths too much to achieve instant efficiency will result in collapsing a natural crowd flow and may eventually put everyone, including the self, at risk of collisions. On the other hand, keeping in silence to avoid every single collision will lead to the agent's inefficient travel. With this observation, our L2B framework augments the reward function used in learning an interactive navigation policy to penalize frequent active path clearing and passive collision avoidance, which substantially improves the balance of the safety-efficiency trade-off. We evaluate our L2B framework in a challenging crowd simulation and demonstrate its superiority, in terms of both navigation success and collision rate, over a state-of-the-art navigation approach.

\end{abstract}

\section{INTRODUCTION}
\begin{figure}[t]
    \centering
    \includegraphics[width=0.93\linewidth]{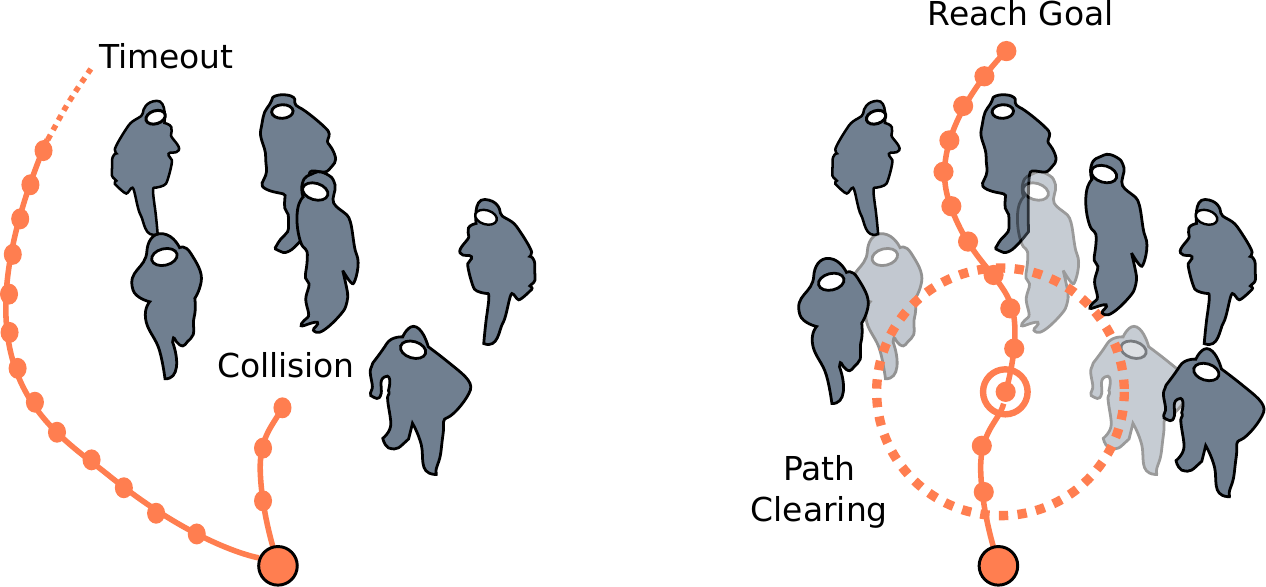}
    \caption{Left: freezing robot problem, where a robot is struggling to find a bypass and as a result, takes unnecessary maneuvers. Right: our agent that can interact with a crowd by actively clearing a path or passively finding a bypath. Our goal is to enable such agents to balance the safety-efficiency trade-off in a highly congested environment.}
    \label{fig:teaser}
\end{figure}

We envision a future mobile robot system that can navigate crowded places, such as busy shopping malls and airports, as naturally as we do. Developing such an intelligent navigation system would enhance several practical applications including automated delivery services \cite{feil2011socially} and guidance at airports \cite{kayukawa2019bbeep}. As shown in \FIGREF{fig:teaser}, to achieve this goal, we present a deep reinforcement learning (RL) framework for crowd-aware navigation, which enables agents to interact with a crowd not only by finding a bypass \emph{safely} but also by actively clearing a path to arrive at their destinations \emph{efficiently}.

Typically, it is easy for humans to navigate in congested environments safely and efficiently. For example, if someone cuts right in front of us, we simply stop walking to avoid potential collisions (\ie, safe navigation). Also, if we are in a hurry to arrive at a destination in time, we typically call out to nearby people to make room (\ie, efficient navigation). However, learning such advanced navigation skills from scratch is non-trivial because there exist no optimal solutions to determine when to avoid collisions passively or to clear a path actively. In the robotics domain, most prior works on mobile robot navigation have only focused on collision avoidance skills~\cite{borenstein1990real,fox1997dynamic,trautman2010unfreezing,chen2017decentralized}. While they allow robots to move safely, too much collision avoidance also results in highly evasive maneuvers~\cite{trautman2010unfreezing,trautman2015robot}. On the other hand, actively addressing nearby people to move away, via sound notifications~\cite{kayukawa2019bbeep} or visual signs~\cite{matsumaru2006mobile,matsumaru2007mobile,watanabe2015communicating}, would allow the robots to navigate efficiently by sticking with the original path plan. However, frequently making room in a crowd could collapse a natural crowd flow and may even increase the risk of collisions, especially in the extremely congested environments. Therefore, there is a trade-off between agent's safety and efficiency in crowd-aware navigation, which begs our key question: \emph{``how can agents learn to balance the safety-efficiency trade-off?''}

To address the above question, we develop a deep RL framework called \emph{Learning to Balance (L2B)}. The proposed L2B framework enables crowd-aware navigation agents such as mobile robots to learn a hybrid policy allowing for choosing to either 1) seek a bypass (to passively avoid potential collisions with a crowd) or 2) actively address nearby persons, \eg, by emitting a beeping sound~\cite{kayukawa2019bbeep} (to clear a planned path for a safe and efficient navigation). Our key insight is that the safety-efficiency trade-off may be viewed as a consequence of social dilemmas between a robot agent and a crowd. Specifically, they can both move safely and efficiently if they mutually cooperate and give way to each other, while doing otherwise will eventually result in navigation inefficiency (\ie, sucker outcome) or unexpected collisions (\ie, punishment from mutual defection). With this insight, we leverage the concept of Sequential Social Dilemmas (SSDs)~\cite{leibo2017multi} to augment the reward function used in learning the navigation policy, where the augmented reward function penalizes the undesirable outcomes and encourages mutual cooperation to balance travel safety and efficiency.

We evaluate the effectiveness of the L2B framework with a simulation for challenging crowd-aware navigation tasks. Our experimental results demonstrated that the L2B outperformed the state-of-the-art SARL~\cite{chen2019crowd} navigation method. To the best of our knowledge, our work is the first to present a deep RL framework for learning a crowd-aware navigation policy that not only avoids collisions passively but also clears the path actively to make agent's travels safe and efficient.

\section{RELATED WORK}
This section reviews related work mainly on 1) crowd-aware robot navigation, 2) modeling of social interactions in navigation tasks, and 3) modeling of crowd dynamics.

\subsection{Crowd-aware Robot Navigation}
A task of navigating robots in a crowd has been addressed based on reciprocal force models~\cite{van2008reciprocal,van2011reciprocal,ferrer2013robot} or imitation learning~\cite{tai2018socially,long2017deep,liu2018map,pfeiffer2016predicting}.
Recently, deep RL approaches have achieved promising results in congested scenarios by jointly performing path planning and collision avoidance~\cite{chen2017decentralized,chen2019crowd,chen2017socially,everett2018motion}.
All of these prior works, however, do not necessarily ensure the efficient travel of robots and can easily get trapped into a ``freezing robot problem''~\cite{trautman2010unfreezing,trautman2015robot}. Specifically, once the environment surpasses a certain level of congestion, the planner decides
that all forward paths are unsafe, and the robot freezes in place or takes unnecessary roundabout ways to avoid collisions.
Despite much progress made to resolve this freezing problem~\cite{fan2019getting,chen2019robot}, it remains hard to enable robots to find a proper path in highly congested scenarios.
Our work mitigates this problem by taking into account the robot's ability to address surrounding pedestrians to actively clear a path.

\subsection{Modeling Social Norms}
Modeling and learning social norms, to allow robots to interact with humans properly, is also an active topic that has been studied in the context of navigation. Some work tried to learn ``social etiquette'' from pedestrian trajectories~\cite{robicquet2016learning}, which is then extended to robot motion planning~\cite{chen2017socially}. Game-theoretic formulations are also studied widely~\cite{claus1998dynamics,tan1993multi,fisac2019hierarchical} and extended to pedestrian modeling~\cite{ma2017forecasting}.
One important study that inspired our work is so-called Sequential Social Dilemmas (SSD)~\cite{leibo2017multi,jaques2019social,lanctot2017unified}, which models how multiple agents cooperate in a complex pay-off structure with a Partially Observable Markov Decision Process (POMDP). Their multi-agent approach is confirmed effective in some Markov games with complex dilemmas such as the well-known Tragedy of the Commons~\cite{hardin1968tragedy}. In this work, we are interested in leveraging their formulation for interactive navigation tasks, where the key novelty is to design a reward function that takes into account the dilemma between a robot agent and a crowd to take a balance between navigation safety and efficiency.

\subsection{Modeling Crowd Dynamics}
Finally, modeling the dynamic behaviors of a crowd is crucial for both tasks of crowd-aware navigation and social interaction under crowded situations. Popular approaches include multi-agent interactions with reciprocal force model~\cite{van2008reciprocal,van2011reciprocal} and imitation learning~\cite{tai2018socially,long2017deep,liu2018map}. These methods have been utilized not only in crowd simulations but also in controlling mobile robots navigating among people~\cite{fraichard2020crowd}. In our experiment, we will make use of the Emotional Reciprocal Velocity Obstacles (ERVO) model~\cite{xu2019crowd} for simulating a crowded environment, where each pedestrian will try to reach designated points or escape to safe places when they feel threatened by robot's interventions.

\section{BACKGROUND} \label{sec:preliminary}
\subsection{Reinforcement Learning for Navigation}
In this work, we consider a task of mobile robot navigation through a crowd, which we formulate as a sequential decision making problem with a POMDP. The robot and the crowd in the environment are regarded as two types of agents, namely \textbf{robot agent} and \textbf{crowd agent}, which are each driven by distinct policies $\pi, \tilde \pi$. Specifically, we regard multiple people in the environment as a single virtual agent to enable the modeling of a social dilemma between the robot and the crowd. This single crowd agent has a policy $\tilde \pi$ that is a fixed and unknown function. This allows us to formulate our problem as a standard crowd-aware navigation (same as \cite{chen2019crowd}), where \emph{only robot agent's policy $\pi$ is trainable, and the crowd agent is modeled as a part of the environment.}

At each time step $t$, the state $\hstate_t$ of the crowd agent is partially observable for the robot agent depending on its field of view and hidden goals of each constituent pedestrian in the crowd. Therefore, we describe the crowd agent's state as a tuple $\hstate_t= \langle\hstate_t^o, \hstate_t^h\rangle$, where $\hstate^o$ and $\hstate^h$ are observable (\eg, locations of the pedestrian closest to the robot) and hidden parts of the $\hstate_t$ respectively. Accordingly, the robot agent observes a joint state $\jstate_t = \langle \rstate_t, \hstate^o_t\rangle$ in each time step, where $\rstate_t$ is its own state such as the location of the self. Then the agent executes an action $\bm a_t$ based on its policy $\pi$, and it will receive an instant reward $R(\jstate_t,\bm a_t)$ designed to encourage the robot when reaching a goal and to penalize collisions with nearby pedestrians. The agent is then transitioned to the next state $\jstate_{t+1}$ based on the hidden state transition of crowd agent from $\langle\hstate_t^o, \hstate_t^h\rangle$ to $\langle\hstate_{t+1}^o, \hstate_{t+1}^h\rangle$.

With reinforcement learning, our objective is to obtain an optimal policy for the robot, $\pi^*:\jstate_t\mapsto\bm a_t$ that maximizes the expectation of discounted total rewards. Following \cite{chen2019crowd}, one can also find the optimal value function $V^*$ that encodes an estimate of the expected return as follows:
\begin{equation}
  \begin{aligned}
    V^*(\jstate_t) = \mathbb E\left[\sum_{t'=t}^{T} \gamma^{t'\cdot \vpref} R\left(\jstate_{t'}, \pi^*(\jstate_{t'})\right) \right],
  \end{aligned}
  \label{eq:optimal-valuefunc}
\end{equation}
 where $\gamma \in [0,1)$ is a discount factor, and preferred velocity $\vpref$ is used to normalize a term in the discount factor for numerical stability~\cite{chen2017decentralized}.
With the value iteration method, the optimal policy $\pi^*(\jstate_t)$ can then be derived as:
\begin{equation}
\begin{aligned}
    \!\!\!\! \pi ^*(\jstate_t) &= \argmax_{\bm a_t} R(\jstate_t, \bm a_t) + \\
    & \gamma ^{\Delta t \cdot \vpref} \int _{\jstate_{t+\Delta t}}\!\!\!\!\!\!\!\!\!\!T(\jstate_t, \bm s^{jn}_{t+\Delta t} |\bm a_t) V^*(\bm s_{t+\Delta t}^{jn})d \jstate_{t+\Delta t}.
\end{aligned}
\end{equation}
To better deal with a high-dimensional state space, we approximate the value function $V$ with a deep neural network.

\subsection{Modeling Social Dilemmas}
As we described earlier, we observe that a robot agent navigating in a crowd involves a type of social dilemmas. In the context of sequential decision making problems, Sequential Social Dilemmas (SSDs) deal with situations where individual agents are tempted to increase their own payoffs at the cost of lowering total rewards~\cite{leibo2017multi}. Specifically, SSDs consider a two-player partially observable Markov game with 
the joint policy denoted by $\jpolicy = (\pi, \tilde \pi)$. 
The long-term payoff $V_i^{\jpolicy}(\rstate_0)$ to player $i$ from $t=0$ can be represented as 
\begin{equation}
  \begin{aligned}
    V_i^{\jpolicy}(\rstate_0) = \mathbb E\left[\sum_{t=0}^{T} \gamma^{t\cdot \vpref} R\left(\rstate_t, \jpolicy (\rstate_t)\right) \right].
  \end{aligned}
\end{equation}
Let $\pi^c$ and $\pi^d$ be cooperative and defecting policies in social dilemma games. In the SSDs, we consider four possible outcomes
$\langle \mathcal R, \mathcal P, \mathcal S, \mathcal T \rangle := \langle V^{\pi^c,\pi^c}(\rstate_0), V^{\pi^d,\pi^d}(\rstate_0), V^{\pi^c,\pi^d}(\rstate_0), V^{\pi^d,\pi^c}(\rstate_0) \rangle$, where $\mathcal{R}$ is \emph{a reward of mutual cooperation}, $\mathcal{P}$ is \emph{a punishment from mutual defection}, $\mathcal{S}$ is \emph{a sucker outcome for cooperation with a defecting partner}, and $\mathcal{T}$ is \emph{a temptation outcome when defecting against a cooperative partner}. They satisfy the following inequalities; a) $\mathcal R > \mathcal P$ (mutual cooperation is better than mutual defection), b) $\mathcal R > \mathcal S$ (and is also better than being exploited), c) $\mathcal T > \mathcal R$ (exploiting the other is preferred to mutual cooperation), and d) $2 \mathcal R > \mathcal T + \mathcal S$ (unilateral cooperation or defecting at equal probability is worse than mutual cooperation). In the next section, we augment our reward function by involving the SSD-like pay-off structure, which allows us to naturally balance the safety-efficiency trade-off in interactive navigation tasks.
\section{APPROACH}
\subsection{Social Dilemmas in Crowd-aware Navigation}\label{sec:navigation-dilemma}
For a robot agent that is capable of clearing a path actively, it is not necessarily optimal to do so whenever it finds someone on the path. Such actions can collapse a natural crowd flow, and may even lead to significant delays in reaching the goal or unexpected collisions if the place around the robot agent gets extremely crowded. On the other hand, keeping silence and avoiding every collision passively, which is what most of the existing approaches do, is inefficient especially in crowded scenes~\cite{chen2019robot}.

In the SSD's terminology, $\pi^c$ can be viewed as a passive policy of one type of agent to avoid collisions by giving way to the other, whereas $\pi^d$ is the contrary. Then, mutual cooperation $\mathcal R=V^{\pi=\pi^c,\tilde\pi=\pi^c}$ holds when robot and crowd agents give way to each other for making each travel reasonably safe and efficient. The robot agent suffers from navigation inefficiency due to frequent passive collision avoidance, which corresponds to $\mathcal S=V^{\pi=\pi^c,\tilde\pi=\pi^d}$. Although the robot agent may get close to the goal faster via active path clearing \ie, $\mathcal T=V^{\pi=\pi^d,\tilde\pi=\pi^c}$, that will eventually result in mutual defection $\mathcal P=V^{\pi=\pi^d,\tilde\pi=\pi^d}$ once the crowd agent is no longer able to make room for the robot in a collapsed flow. 

\subsection{Learning to Balance the Safety-Efficiency Trade-off}
\label{sec:ssd-reward}

With the above insight, we propose to better balance the navigation safety and efficiency by encouraging the robot to stay in $\mathcal{R}$ while avoiding $\mathcal{S}$ and $\mathcal{T}$ that leads to $\mathcal{P}$. To incorporate this SSD structure into RL-based navigation tasks, we augment the reward function $R(\jstate_t, \bm a_t)$ to take into account the result of the robot's active path clearing:
\begin{align}
  \label{eq:ssd-reward}
  R(\jstate_t,\bm a_t) = R\sub{e} (\jstate_t, \bm a_t) + R\sub{s} (\jstate_t, \bm a_t),
\end{align}
where $R_e$ is the reward from environments, and $R_s$ is that from the crowd agent. $R_e$ is given by:
\begin{align}
R\sub{e}(\jstate_t, \bm a_t) = \begin{cases}
  1.0 - \alpha \frac{t}{t\sub{lim}} & \mathrm{if} \, \bm p_t = \bm p\sup{g} \\
  -0.25 & \mathrm{else if}\,d_t < d_{\mathrm{min}}\\
  0 & \mathrm{otherwise},
\end{cases}
\end{align}
where $d_{t}$ is the distance between the robot agent and the crowd agent (\eg, the closest pedestrian) at time $t$ and $t\sub{lim}$ is a time limit for the task completion. $\bm p_t$ and $\bm p\sup{g}$ are the position of the robot at time $t$ and its destination, respectively. This reward will monotonically decay over time so that the robot agent will be encouraged to reach the goal as early as possible.

On the other hand, We define $R_s$ as follows: 
\begin{align}
R\sub{s}(\jstate_t, \bm a_t) = \begin{cases}
  \beta (d_t - r\sup{b}) \!\!& \mathrm{if} \, d_t < r\sup{b} \land b_t=1\\
  \eta (d_{\mathrm{t}}-d_{\mathrm{disc}})  \!\!& \mathrm{else if}\, d_{\mathrm{t}} < d_{\mathrm{disc}}\\
  0 & \text{otherwise},
\end{cases}
\end{align}
where $r\sup{b}$ is the effective range of active path clearing actions within which persons in the crowd will get influenced and move away, and $d\sub{disc}$ is the minimum discomfort distance between the robot agent and the crowd agent set to encourage earlier collision avoidance. \textcolor{\fixcolor}{A binary vector $b_t\in\{0,1\}$ is 1 when path clearing action is invoked by a robot and 0 otherwise.} $\beta$ and $\eta$ are hyper-parameters that respectively adjust the influence of active path clearing and a penalty due to crowd agent's discomfort. \textcolor{\fixcolor}{The first term with $\beta$ represents the robot's aggressiveness in path clearing strategy, where a robot agent is tempted to clear a path rather than finding a bypass, \ie, the transition from $\mathcal{R}$ to $\mathcal{T}$. However, too much active path clearing will harm the crowd flow, \ie, the transition from $\mathcal{T}$ to $\mathcal{P}$.}
On the other hand, the second term with $\eta$ corresponds to $\mathcal{R}$ to $\mathcal{S}$; permitting crowd agent's free movements via collision avoidance too much will make the robot travel inefficient. Moreover, we set $\beta$ and $\eta$ to satisfy $\eta > \beta$; the robot agent will be penalized more heavily if it gets too close to nearby persons than actively clearing a path, \ie, $\mathcal{T}>\mathcal{S}$. A high return by $\mathcal{R}$ is obtained when the robot agent reaches its destination as early as possible to increase $R\sub{e}$ while avoiding $\mathcal{T}, \mathcal{S}, \mathcal{P}$ by keeping the effect from $R\sub{s}$ as small as possible. Doing so corresponds to reasonably reducing the frequency of both crowd avoidance and active path clearing; \ie, balancing the safety-efficiency trade-off.

With this reward function, the value function can be trained via a standard V-learning based on a temporal difference method with experience replay and fixed target techniques~\cite{mnih2015human}, such as done in \cite{chen2019crowd}.
\section{SIMULATION}
To evaluate the effectiveness of the proposed L2B framework, we develop a simulation environment on top of the OpenAI Gym~\cite{openaigym} for challenging interactive crowd-aware navigation tasks.

\subsection{Environment Setup}
Following \cite{chen2017decentralized,chen2019crowd}, we implement circle crossing scenarios where $N$ pedestrians and the robot are positioned randomly on a circle with a certain fixed radius, which will each walk to their destination set at the opposite side of the circle (see also Fig.~\ref{fig:qual}). Random noises are added to the initial locations and destinations to make scenes further diverse.

For the $i$-th pedestrian, let $\bm p\supi{i}_t \in\mathbb{R}^2$ and $\bm v\supi{i}_t \in\mathbb{R}^2$ be the location and velocity at time $t$. Similarly, let $\bm p_t\in\mathbb{R}^2$ and $\bm v_t\in\mathbb{R}^2$ be those of the robot agent. We also denote the sizes of the pedestrian and the robot by $r\sup{i}, r\sup{c}\in\mathbb{R}_+$, and the distance between the $i$-th pedestrian and the robot at $t$ by $d\supi{i}_t=\|\bm p\supi{i}_t - \bm p_t\|\in\mathbb{R}_+$. They are judged to collide with each other if $r\sup{c} + r\sup{i} \geq d\supi{i}_t$. As for the robot's ability to actively clear its own path, let $b_t\in\{0, 1\}$ be a binary vector indicating if an active path clearing action (which we also refer to as a \emph{beep} action in our experiment section by following~\cite{kayukawa2019bbeep}) is executed at time $t$, and $r\sup{b}\in\mathbb{R}_+$ be its effective range. Finally, let $\bm p\sup{g}$ be the robot's destination, $d\sup{g}_t=\|\bm p_t - \bm p\sup{g}\|$ be the direct distance from the current robot's location to the destination, and $v\sub{pref}\in\mathbb{R}_+$ be a preferred travel speed of the robot.

With the above notations, a state of the robot agent $\rstate_t$ and that of the crowd agent $\hstate^o_t$ are defined as follows:
\begin{align}
  \rstate_t &= [d\sup{g}_t, \bm v_t, \vpref, r\sup{c}, r\sup{b}], \\
  \hstate^o_t &= [\tilde{\bm p}_t, \tilde{\bm v}_t, \tilde d_t, \tilde r],
\end{align}
where $\tilde{\bm p}_t =\bm p^{(j)}_t$, $j = \arg\min_i d\supi{i}_t$, is the location of pedestrian located closest to the robot at time $t$, and the same applies to $\tilde{\bm v}_t, \tilde d_t, \tilde r$. Also, the robot agent's action $\bm a_t$ is defined using  $\vct{v}_{t+1},b_{t+1}$ (will be given concretely in the next section), assuming that the velocity of the robot can be controlled instantly.

\subsection{Modeling Reactions of Pedestrians}
\label{sec:reactive-agent}

One critical choice of designs for simulating interactive navigation scenarios is how a crowd reacts to active path clearing by the robot. In this work, we implement a simplified version of Emotional Reciprocal Velocity Obstacles (ERVO) \cite{xu2019crowd}, which is an extension of RVO considering a person's emotional reaction towards a threat. ERVO simulates how people in a panic choose their paths to safe places or a planned goal in a realistic way. 

\begin{figure}[t]
    \centering
    \includegraphics[width=0.88\linewidth]{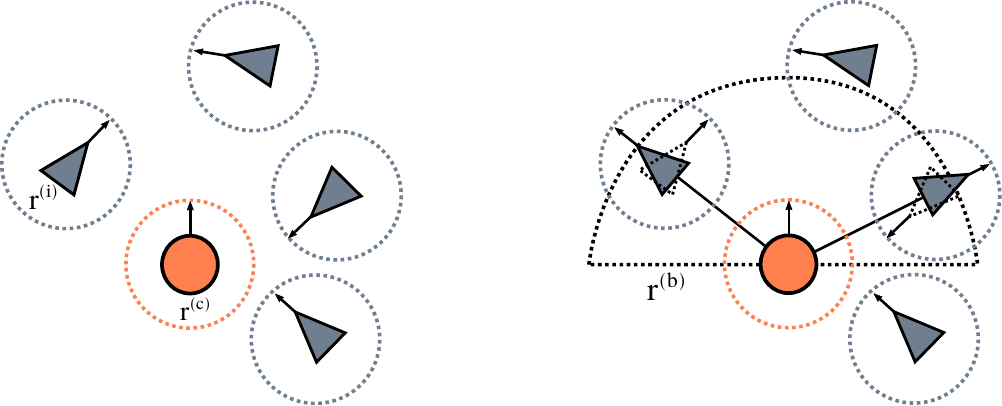}
    \caption{ERVO-based reactive pedestrians.
    A circle and triangles represent a robot with size $r\sup{c}$ and pedestrians with size $r\sup{i}$, respectively (Left). When the robot executes an active path clearing action, all the pedestrians within the half-circle with radius $r\sup{b}$ (Right) are affected to change their directions based on \EQREF{eq:ervo-eq}.}
    \label{fig:ervo}
\end{figure}

Formally, let $\Gamma({\bm p}\supi{i}_t)$ be the degree of influence of active path clearing for a pedestrian at location ${\bm p}\supi{i}_t$. We represent this influence with a Gaussian distribution such as
\begin{align}
\!\!\!\!\!\Gamma ({\bm p}\supi{i}_t)\!=\!\! \begin{cases}
\frac{1}{\sqrt{2\pi}r\sup{b}}\exp(-\frac{({\bm p}\supi{i}_t-\bm p_t)^2}{{2r\sup{b}}^2})& \!\!\!\!\!\mathrm{if}\, d\supi{i}_t \!\!\!<\! r\sup{b} \!\!\land \! b_t\!=\!1\\  0 & \!\!\!\!\mathrm{otherwise}.
\end{cases}
\end{align}
Then, the $i$-th agent will change its velocity $\bm v\supi{i}_t$ as follows:
\begin{align}
  \bm v\supi{i}_t \leftarrow \begin{cases}
    \Gamma(\bm p\supi{i}_t) \cdot \frac{\bm p\sup{i}_t-\bm p_t}{d\supi{i}_t} & \!\!\!\mathrm{if}\, d\supi{i}_t <r\sup{b} \land b_t =1\\ 
    \bm v\supi{i}_t & \!\!\!\mathrm{otherwise.}
  \end{cases}
  \label{eq:ervo-eq}
\end{align}
As illustrated in \FIGREF{fig:ervo}, the agents within $r\sub{b}$ will act to escape from the path clearing influence that attenuates based on the Gaussian distribution.
On the basis of these fundamental principles of ERVO, we extend a collision avoidance simulation called Optimal Reciprocal Collision Avoidance (ORCA)~\cite{van2011reciprocal} to synthesize realistic flows of a crowd.

\section{EVALUATIONS} 
\def\figcolwidth{0.24}
\begin{figure*}[t]
  \centering
  \begin{minipage}[t]{\figcolwidth\linewidth}
    \centering
    \includegraphics[width=1.0\linewidth]{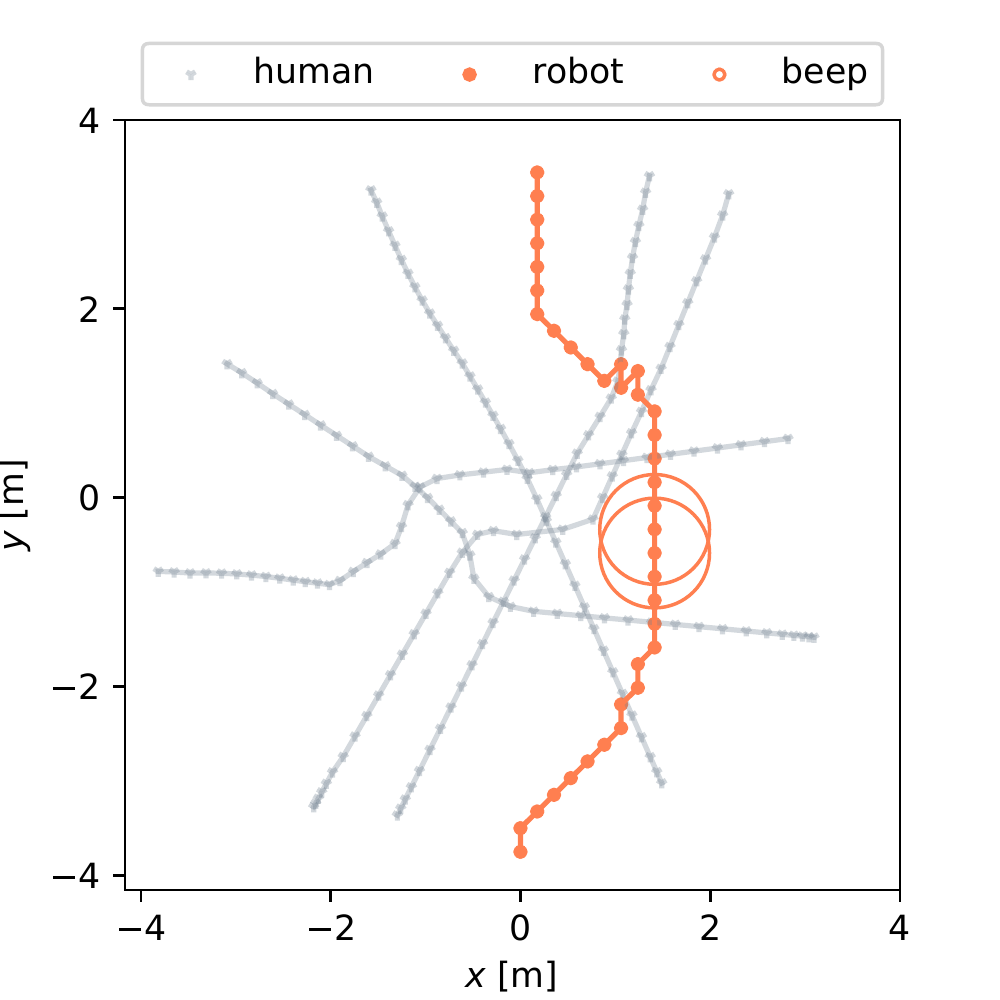} 
    \subcaption{\proposed~($N=5$)}
  \end{minipage} 
  \begin{minipage}[t]{\figcolwidth\linewidth}
     \centering
     \includegraphics[width=1.0\linewidth]{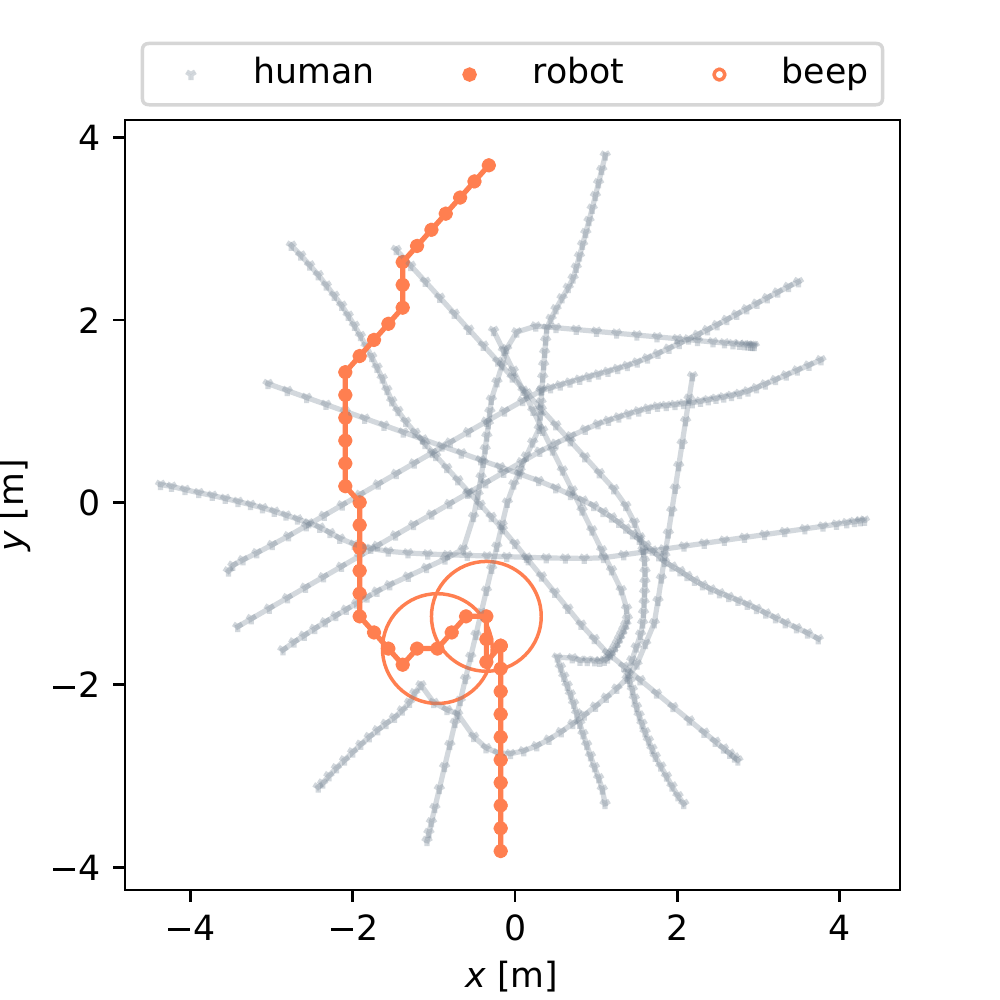}
     \subcaption{\proposed~($N=10$)}
   \end{minipage} 
   \begin{minipage}[t]{\figcolwidth\linewidth}
    \centering
    \includegraphics[width=1.0\linewidth]{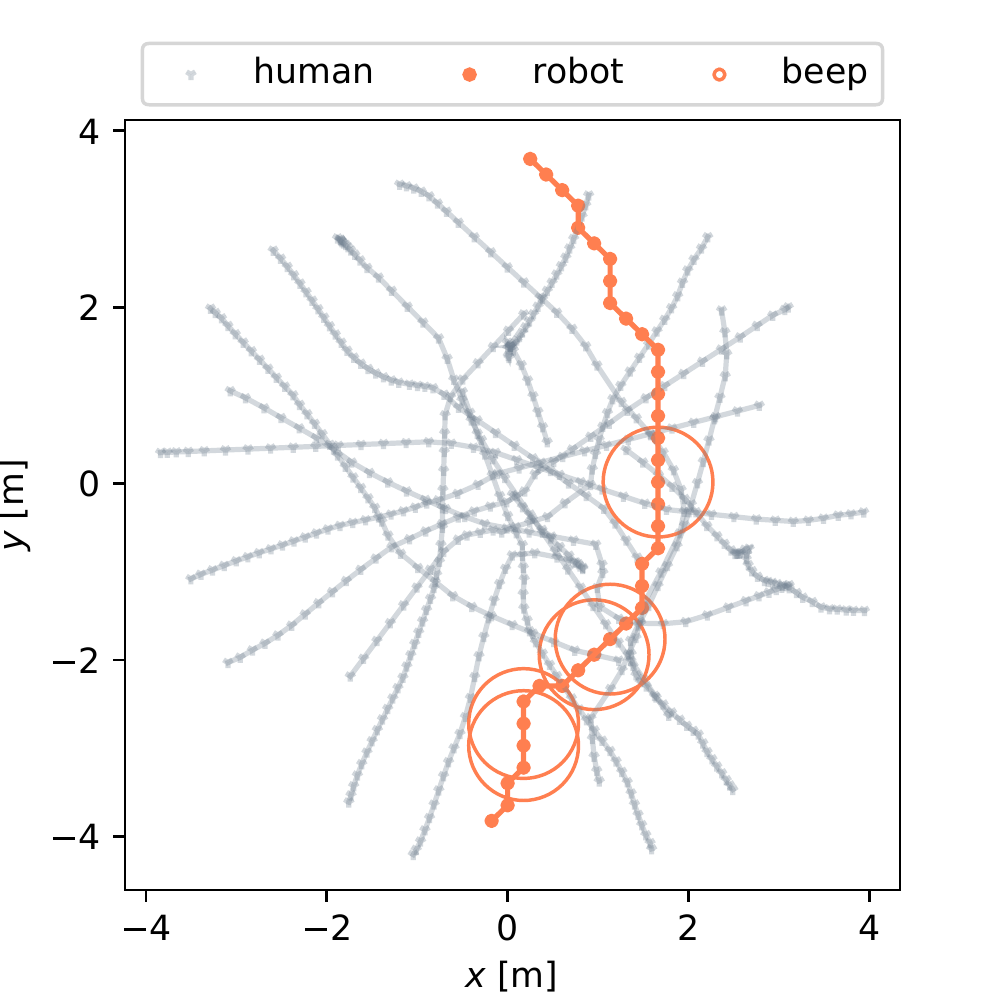} 
    \subcaption{\proposed~($N=15$)}
  \end{minipage} 
  \begin{minipage}[t]{\figcolwidth\linewidth}
    \centering
    \includegraphics[width=1.0\linewidth]{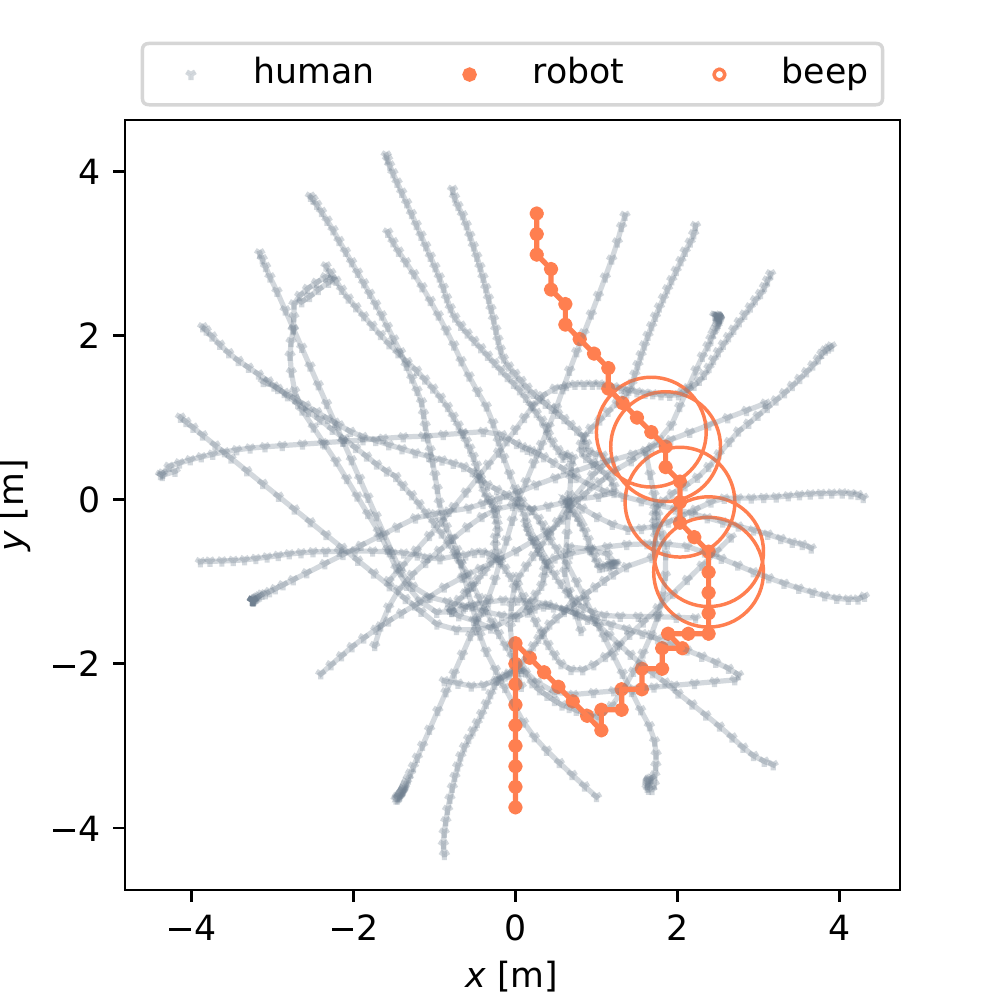} 
    \subcaption{\proposed~($N=20$)}
  \end{minipage} 
    \begin{minipage}[t]{\figcolwidth\linewidth}
    \centering
    \includegraphics[width=1.0\linewidth]{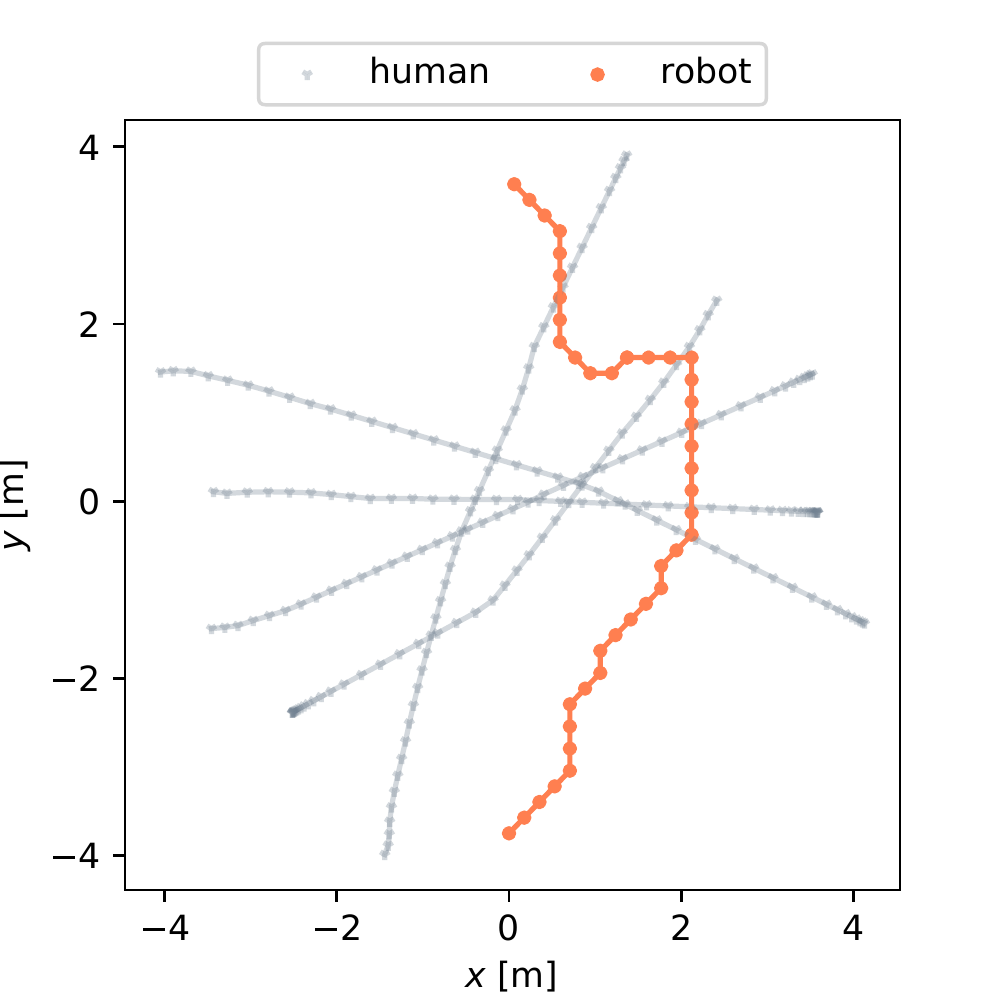} 
    \subcaption{SARL~($N=5$)}
  \end{minipage} 
      \begin{minipage}[t]{\figcolwidth\linewidth}
    \centering
    \includegraphics[width=1.0\linewidth]{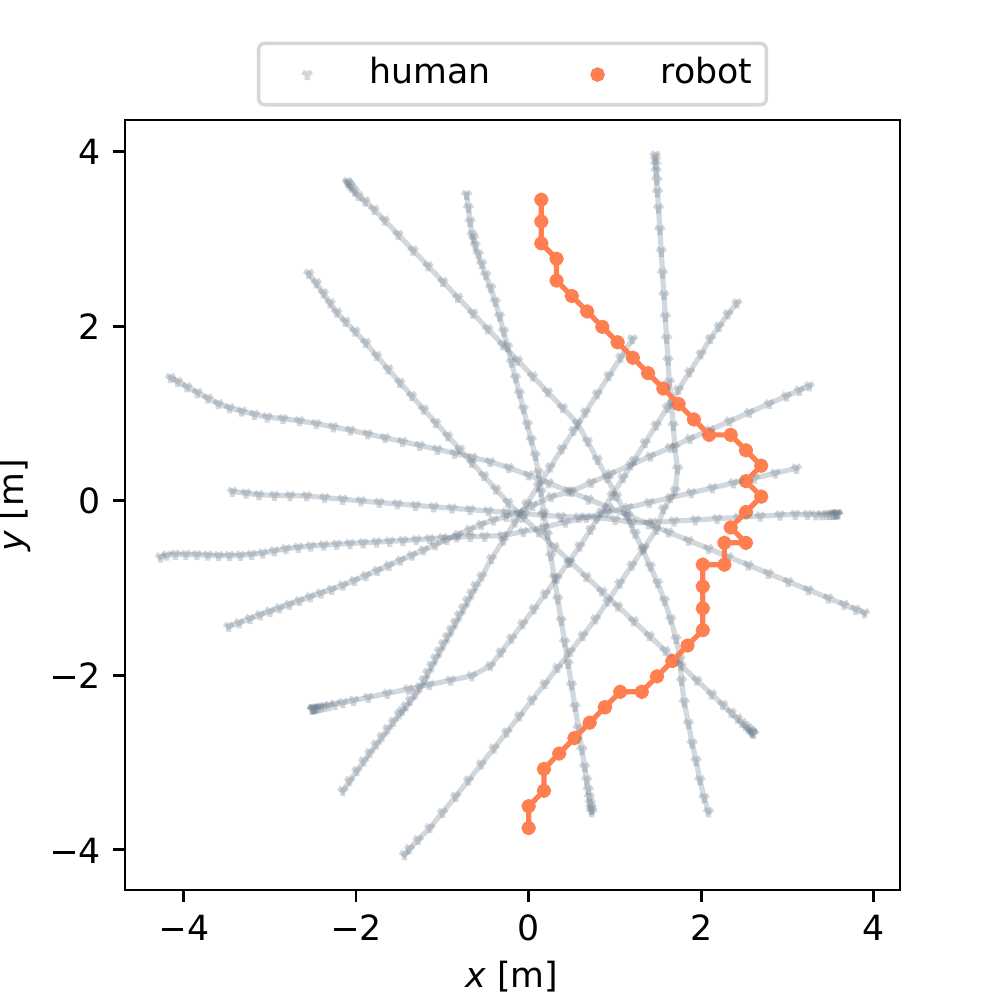} 
    \subcaption{SARL~($N=10$)}
  \end{minipage} 
  \begin{minipage}[t]{\figcolwidth\linewidth}
    \centering
    \includegraphics[width=1.0\linewidth]{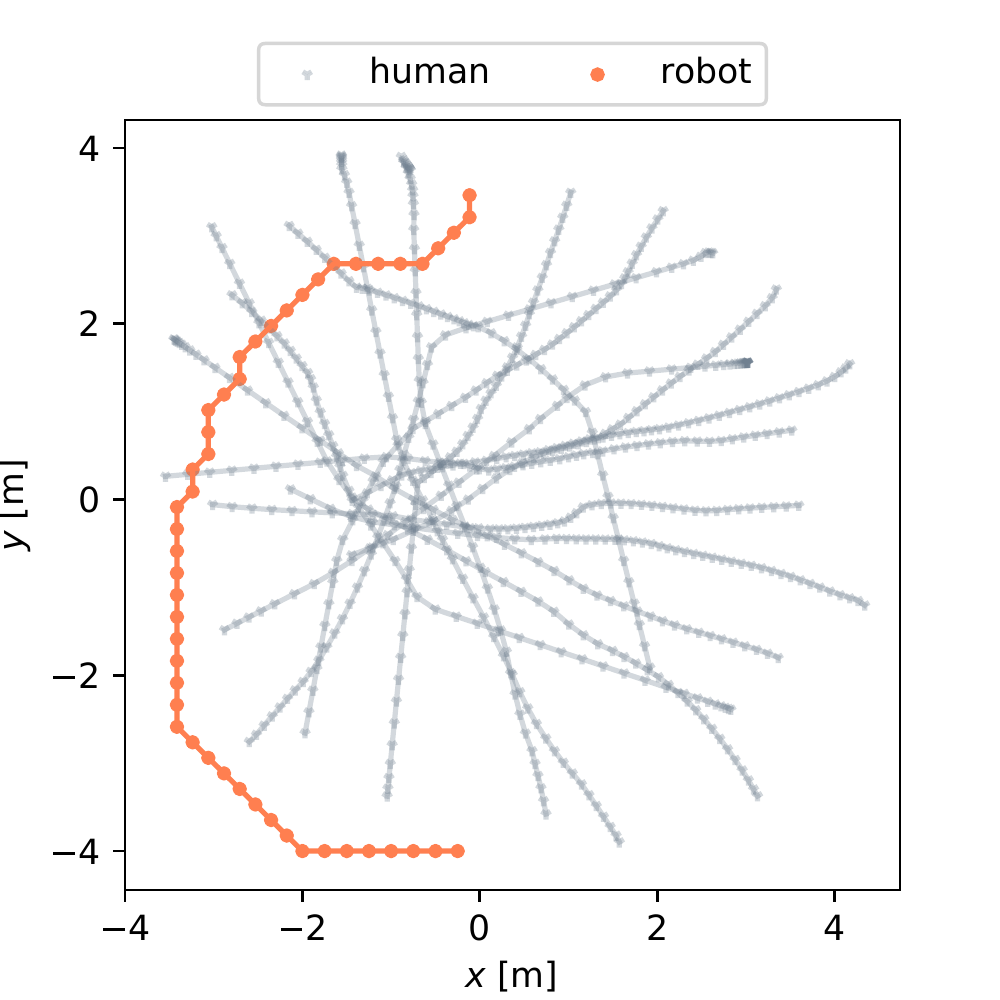} 
    \subcaption{SARL~($N=15$)}
  \end{minipage} 
  \begin{minipage}[t]{\figcolwidth\linewidth}
    \centering
    \includegraphics[width=1.0\linewidth]{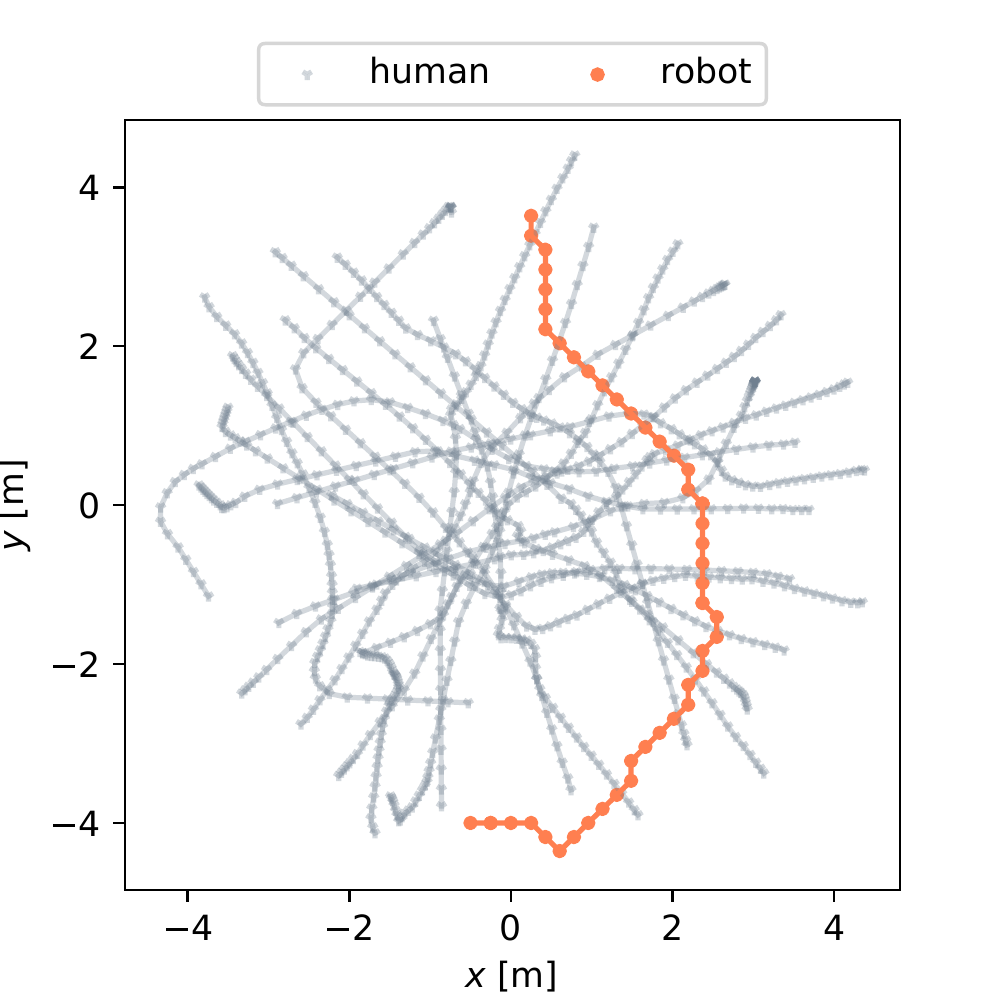} 
    \subcaption{SARL~($N=20$)}
  \end{minipage} 

 \caption{Qualitative results. (a)-(d) show results of our proposed method \proposed , and (e)-(h) show those of SARL. Gray points represent human agents, and filled orange circle represents robot agents. Bigger orange circle represent path clearing beep actions.}
 \label{fig:qual} 
\end{figure*}

\def\figcolwidth{0.49}
\begin{figure}[t]
  \centering
  \begin{minipage}[t]{\figcolwidth\linewidth}
    \centering
    \includegraphics[width=1.0\linewidth]{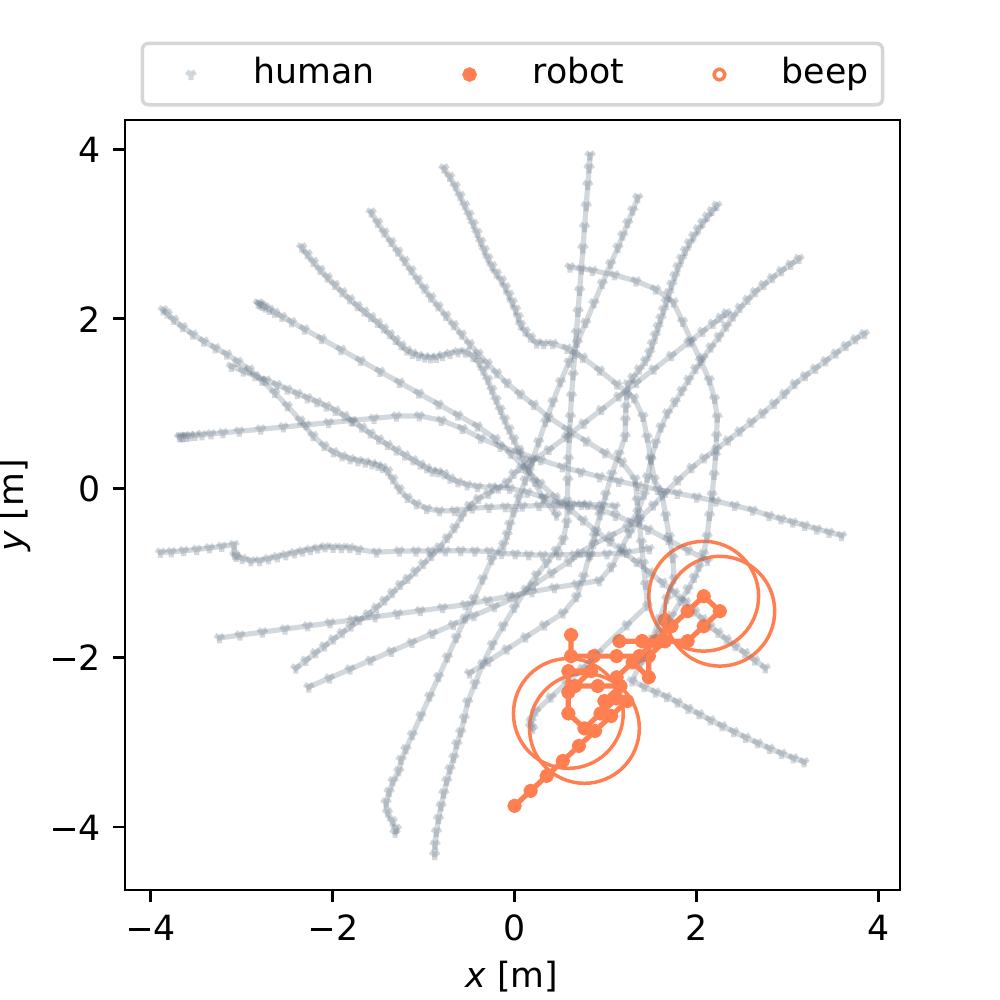} 
    \subcaption{APC frequency: high}
  \end{minipage} 
  \begin{minipage}[t]{\figcolwidth\linewidth}
     \centering
     \includegraphics[width=1.0\linewidth]{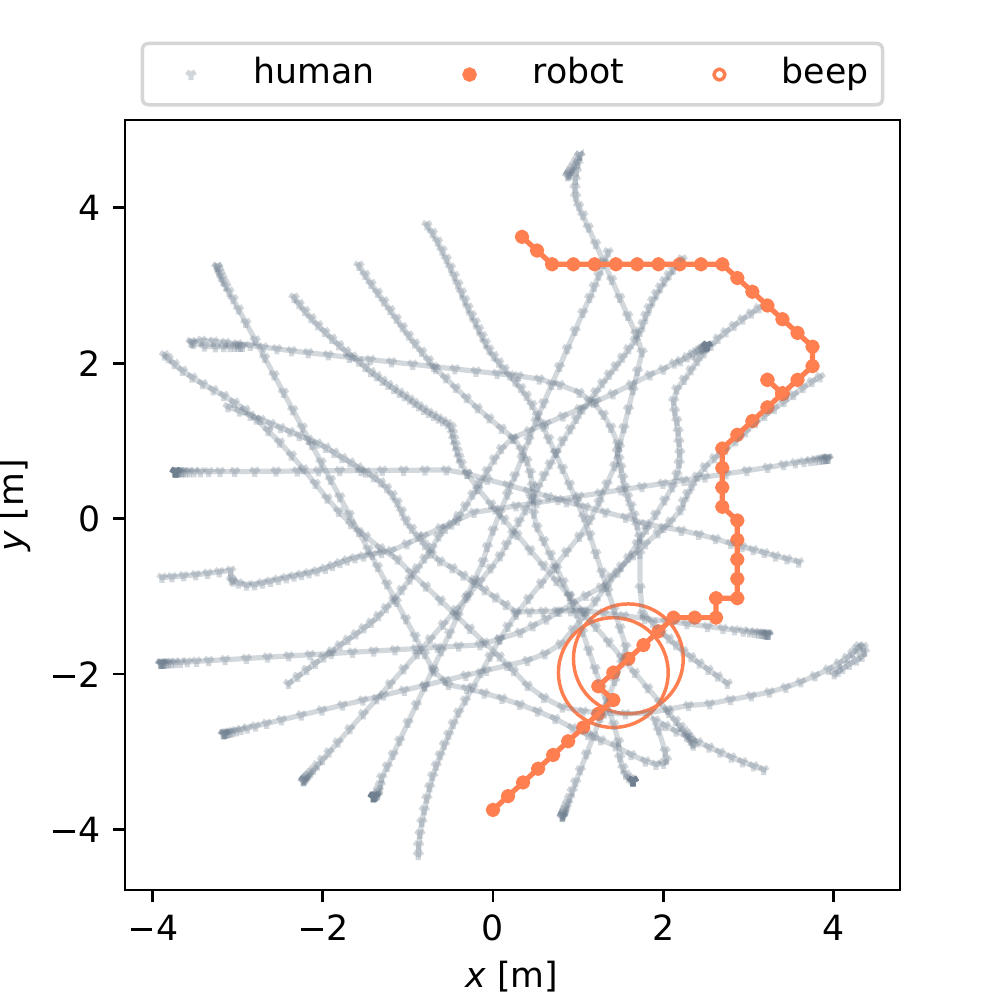} 
     \subcaption{APC frequency: balanced}
   \end{minipage} 
 \caption{Effect of path clearing frequency. (a) When $\beta$ is set to be small, too many active path clearing (APC) caused traffic confusions and increased the possibility of collisions. (b) Our approach can learn to choose path clearing and collision avoidance adequately.}
 \label{fig:qual2} 
\end{figure}

\def\arraystrechlen{1.6}
\begin{table}[t]
\caption{Quantitative results. Success rate, collision rate, timeout rate, and average navigation time for the proposed L2B-SARL and the baseline SARL under different numbers of pedestrian $N$ in a crowd.} 
  \renewcommand{\arraystretch}{\arraystrechlen}
  \centering
  \begin{tabularx}{0.98\linewidth}{cccccc}
    \toprule[1pt]
     Methods & $N$ & Success & Collision & Timeout & Time\\
    \midrule
        \textbf{\proposed} &\multirow{2}{*}{20}&\textbf{0.906}&\textbf{0.094}& \textbf{0.000} & 13.85 \\
        SARL~\cite{chen2019crowd}&  & 0.700 & 0.184 & 0.116 & \textbf{13.50} \\
    \midrule
        \textbf{\proposed} & \multirow{2}{*}{15} & \textbf{0.880} & 0.118 & \textbf{0.002} & \textbf{11.60}\\
        SARL~\cite{chen2019crowd}& & 0.778 & \textbf{0.064} & 0.158 & 12.43\\
    \midrule
        \textbf{\proposed} &\multirow{2}{*}{10}& 0.904 & 0.086 & \textbf{0.004} & \textbf{11.31} \\
        SARL~\cite{chen2019crowd}&  & \textbf{0.922} & \textbf{0.046} & 0.032 & 11.91 \\
    \midrule
        \textbf{\proposed}  & \multirow{2}{*}{5} & \textbf{0.978} & \textbf{0.020} & 0.002 & 10.14\\
        SARL~\cite{chen2019crowd}&  & 0.966 & 
    0.032 & 0.002 & \textbf{10.09}\\
    \bottomrule
        \textbf{\proposed}  & \multirow{2}{*}{average}  & \textbf{0.917} & \textbf{0.079} & \textbf{0.002} & \textbf{11.72}\\
        SARL~\cite{chen2019crowd}&  & 0.841 &  0.081 & 0.077 & 11.98\\
    \bottomrule
  \end{tabularx}
\label{table:quant}
\end{table}

With the simulation introduced above, we evaluate a performance of a state-of-the-art crowd-aware navigation method called \textbf{SARL}~\cite{chen2019crowd}, and its extended version equipped with the active path clearing ability trained in the proposed L2B framework, which we will refer to as \textbf{L2B-SARL}.

\subsection{Experimental Setup}
\subsubsection{Environments}
With our simulation, we synthesize diverse crowd-aware navigation tasks with the number of pedestrians $N\in\{5, 10, 15, 20\}$. Kinematics of the robot agent is assumed to be holonomic, \ie, it can move in any direction without spin. The velocity of robot, $\bm v_t$, was discretized in two speeds in $\{0,v\sub{pref}\}$ and eight orientations spaced evenly between $[0,2\pi)$. In total, there were 9-dimensional action spaces for SARL (\ie, moving one of the eight directions with $v\sub{pref}$ or standing still), and 17-dimensional action spaces (eight directions $\times$ with/without path-clearing beeps, and standing still) for the L2B-SARL.
The preffered speed of the agent $v\sub{pref}$ was set to $1.0$m/s.
For the proposed reward function in Section~\ref{sec:ssd-reward}, we set discomfort distance $d_{\mathrm{disc}}$ to be $0.2$m, the effective range of path clearing actions $r\sup{b}$ to be $1.0$m, and the diameter of both the robot and the human agents $r\sup{c}$, $r\sup{i}$ to be $0.3$m.
The attenuating reward coefficient for reaching a goal $\alpha$ and discomfort penalty factor $\eta$ are configured to be $0.1$ and $0.5$, respectively.
For $\beta$, we set $\beta=0.2$ unless specified otherwise.
In the simulation, the radius of circle on which pedestrians and the robot were placed, was set to $4.0$m.

\subsubsection{Training Details}
For both L2B-SARL and SARL, we implemented a value function with the attention-based network proposed in \cite{chen2019crowd}, which allows agents to observe nearby pedestrians effectively. Intuitively, the network models pairwise human-robot interactions explicitly while encoding human-human interactions in a coarse-grained feature map, and aggregates the interactions by a self-attention mechanism. This value network was first initialized via imitation learning from 3k episodes collected from the ORCA~\cite{van2011reciprocal} policy, using the Adam optimizer~\cite{kingma2014adam} with the learning rate of $0.01$ for 50 epochs. Then, we train the value network in an RL loop via the Adam optimizer with learning rate $0.001$, mini-batch of size $100$, and discount factor $\gamma=0.9$, for 20k episodes\footnote{Note that SARL performed poorly in environments with $N=20$, which reached only $0.04$ success rate. Therefore firstly we trained the policy in environments with $N=10$ for the first $10k$ episodes and then transferred to $N=20$ environments for the remaining $10k$ episodes.}. We adopted a standard $\epsilon$-greedy exploration scheme where $\epsilon$ decayed linearly from $0.5$ to $0.1$ in the first 5k episodes and stayed at $0.1$ for the rest.

\subsubsection{Evaluation Scheme}
For each environment with the number of people in crowd $N \in \{5,10,15,20\}$,
we evaluated the trained models with 500 random test cases with the following metrics; ``Success'': the rate of robot reaching its goal without a collision, ``Collision'': the rate of robot terminating its navigation due to collisions with other humans, ``Timeout'': the rate of robot unable to reach a goal within time limit $t_{\mathrm{lim}}$, and ``Time'': the average robot’s navigation time to reach its goal in seconds.
For each evaluation, the random seed was set to be the same so that each test case can be evaluated on exactly the same sequence.

\subsection{Results}
\subsubsection{Quantitative Evaluation}
Table \ref{table:quant} summarizes quantitative evaluation results. Overall, we confirmed that L2B-SARL demonstrated high success rates regardless of the number of people $N$, whereas the baseline SARL degraded its performance as the environment got more crowded. This is mainly because the baseline agent could only find a bypass passively when it found someone on its path, resulting in high timeout rates. On the other hand, the L2B-SARL provides an extremely low timeout rate thanks to its ability to actively clear a path, at the small cost of small increases of collision rates. The average navigation time for the successful sequences was comparable between L2B-SARL and SARL, These results show that the proposed L2B framework allows us to better balance the safety and efficiency trade-off in crowd-aware navigation tasks.

\def\figcolwidth{0.49}
\begin{figure}[t]
  \centering
  \begin{minipage}[t]{\figcolwidth\linewidth}
    \centering
    \includegraphics[width=1.0\linewidth]{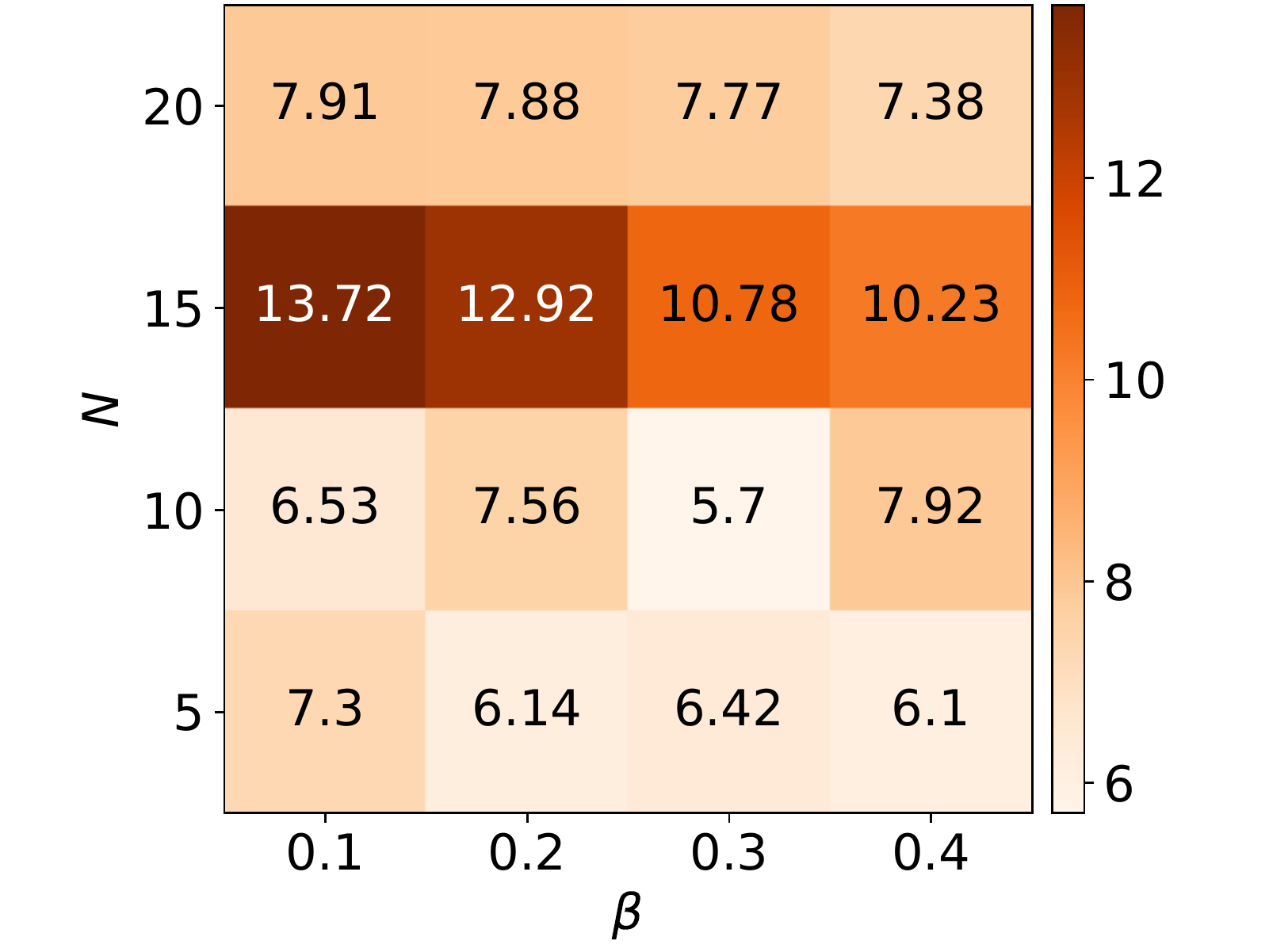} 
    \subcaption{Path clearing frequency [\%]}
  \end{minipage} 
  \begin{minipage}[t]{\figcolwidth\linewidth}
     \centering
     \includegraphics[width=1.0\linewidth]{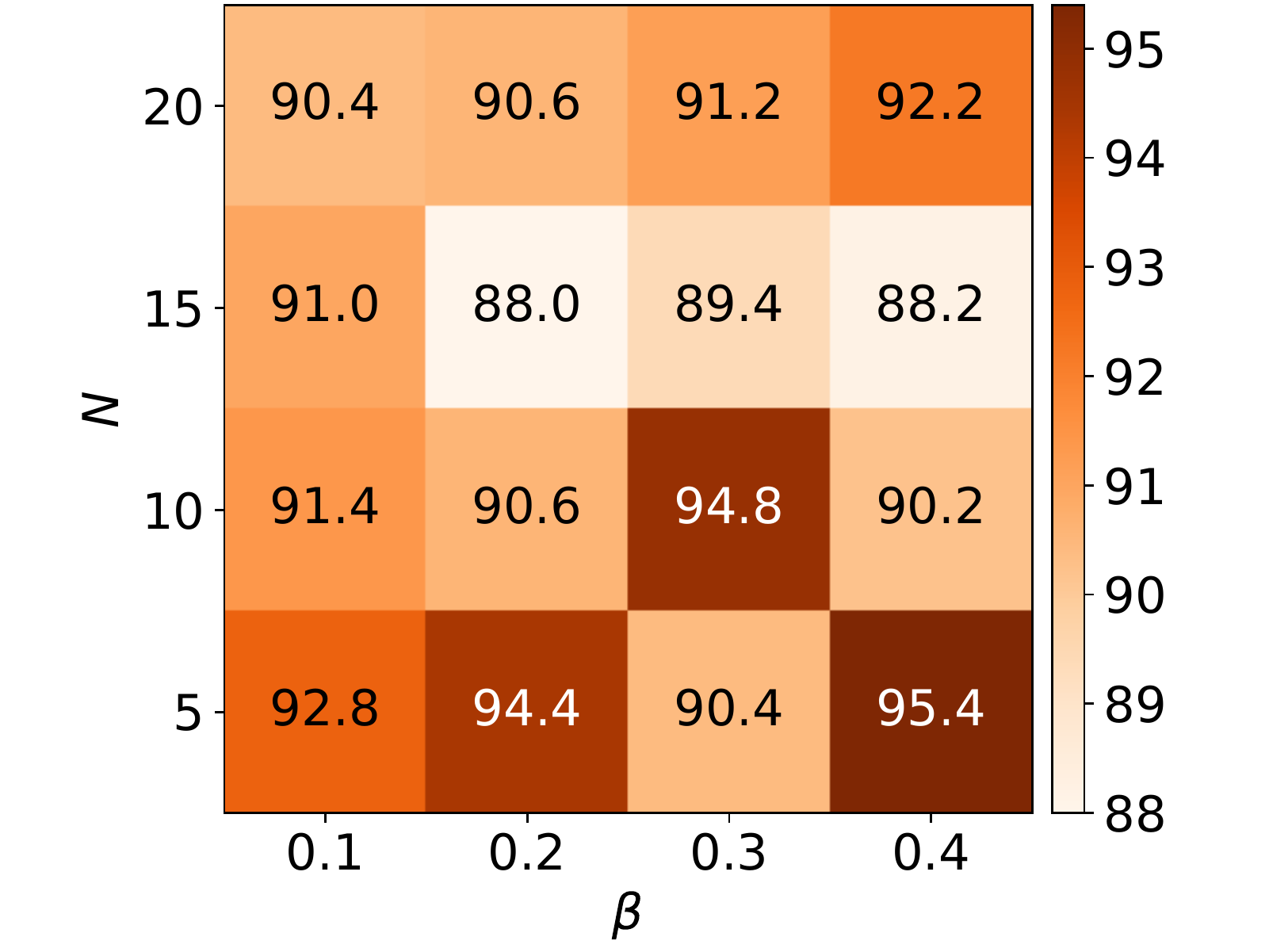} 
     \subcaption{Success rate [\%]}
    \hspace{2em}
   \end{minipage} 
 \caption{Effect of $\beta$. (a) Frequency of active path-clearing actions and (b) success rate of episodes for different combination of penalty factor $\beta$ and the number of pedestrians in a crowd $N$.}
 \label{fig:agg} 
\end{figure}
\subsubsection{Effect of Active Path Clearing}
We further investigated how the robot learned to actively clear a path adequately under several different choices of $\beta$. \FIGREF{fig:agg} summarizes frequencies of path clearing actions (a) and success rates (b) for $\beta\in\{0.1, 0.2, 0.3, 0.4\}$. 
\textcolor{\fixcolor}{Importantly, there was no monotonic tendencies for the path clearing frequency for any choice of $\beta$. 
The robot learned to actively clear a path more frequently as $N$ becomes larger at $N=5\dots 15$, but in highly congested scenarios at $N=20$, kept its frequency lower. 
With constantly high success rates shown in (b), these results indicate that
the ideal path clearing frequency would depend on a congestion factor. More specifically, for less congested environments at $N=5,10$, the frequency is preferred to be low because there are enough spaces to find a bypass as shown in \FIGREF{fig:qual} (a)(e). In more congested environments at $N=15$, the success rate increases as $\beta$ becomes lower since
frequent path clearing allows agents to reach a goal with a shorter path as shown in \FIGREF{fig:qual} (c)(g).
However, for highly congested environments at $N=20$, the frequency is preferred to be low again and the success rate decreases as $\beta$ becomes lower.
One possible interpretation of this result is, in highly congested scenarios as in \FIGREF{fig:qual} (d), too much active path clearing will also make its travel inefficient or unsafe in the presence of social dilemmas against the crowd.
}

\subsection{Qualitative Evaluation}
\FIGREF{fig:qual} visualizes some typical results for L2B-SARL in (a)-(d) and SARL in (e)-(h). We observe that SARL's paths became roundabout and inefficient by trying to bypass crowded regions at the center of the environment in $N=15, 20$. In contrast, the L2B-SARL agent was able to navigate through a crowd by actively clearing its path (denoted by circles). Moreover, \FIGREF{fig:qual2} (a) \textcolor{\fixcolor}{shows the special case that the agent is tempted to clear a path frequently and the episode ended with collision. It indicates that too much path clearing does harm a robot's efficient travel, as doing so is not effective when pedestrians being addressed had little room to move away. If the path clearing frequency is balanced by our learnt adaptive policy, the agent can efficiently reach a goal with the moderate number of times of path clearing as shown in \FIGREF{fig:qual2}(b).}
\section{CONCLUSION}
We have presented a new deep RL framework called L2B for crowd-aware navigation tasks, which enabled robotic agents to navigate through a crowd safely and efficiently. The key idea is to equip the agents with the ability to actively clear a path as well as passively find a bypath to avoid collisions. With a reward function that takes into account the presence of social dilemmas between the robot and a crowd, the proposed L2B framework allows us to learn a navigation policy to choose these two actions adequately to take a good balance between travel safety and efficiency. Our extensive simulation experiments demonstrated the superiority of the proposed approach over a state-of-the-art navigation method.

Currently, we limit our study to assume that all the pedestrians only react passively based on the fixed policy. One interesting extension of the proposed work is to involve pedestrians who also try to clear a path actively. That will make more explicit the presence of a social dilemma structure in a crowd flow, also requiring a new technique for simulating such active pedestrians in a realistic scenario. Another possible direction for future work is to formulate this crowd-aware navigation task in a multi-agent RL problem, where each pedestrian in a crowd also allowed to improve its policy to better cooperate with the robot. Such a direction is beneficial for practical robotics applications such as swarm robotics~\cite{brambilla2013swarm} and multiple vehicle control~\cite{cao2010distributed}.

\section*{Acknowledgments}
The authors would like to thank Mohammadamin Barekatain, Tatsunori Taniai and Yoshihisa Ijiri for the insightful discussions and helpful feedback on the manuscript.

\balance

\bibliographystyle{IEEEtran} 
\bibliography{root}

\begin{thebibliography}{10}
\providecommand{\url}[1]{#1}
\csname url@samestyle\endcsname
\providecommand{\newblock}{\relax}
\providecommand{\bibinfo}[2]{#2}
\providecommand{\BIBentrySTDinterwordspacing}{\spaceskip=0pt\relax}
\providecommand{\BIBentryALTinterwordstretchfactor}{4}
\providecommand{\BIBentryALTinterwordspacing}{\spaceskip=\fontdimen2\font plus
\BIBentryALTinterwordstretchfactor\fontdimen3\font minus
  \fontdimen4\font\relax}
\providecommand{\BIBforeignlanguage}[2]{{%
\expandafter\ifx\csname l@#1\endcsname\relax
\typeout{** WARNING: IEEEtran.bst: No hyphenation pattern has been}%
\typeout{** loaded for the language `#1'. Using the pattern for}%
\typeout{** the default language instead.}%
\else
\language=\csname l@#1\endcsname
\fi
#2}}
\providecommand{\BIBdecl}{\relax}
\BIBdecl

\bibitem{feil2011socially}
D.~Feil-Seifer and M.~J. Matari{\'c}, ``Socially assistive robotics,''
  \emph{IEEE Robotics \& Automation Magazine}, vol.~18, no.~1, pp. 24--31,
  2011.

\bibitem{kayukawa2019bbeep}
S.~Kayukawa, K.~Higuchi, J.~Guerreiro, S.~Morishima, Y.~Sato, K.~Kitani, and
  C.~Asakawa, ``Bbeep: A sonic collision avoidance system for blind travellers
  and nearby pedestrians,'' in \emph{CHI Conference on Human Factors in
  Computing Systems}, 2019, pp. 1--12.

\bibitem{borenstein1990real}
J.~Borenstein and Y.~Koren, ``Real-time obstacle avoidance for fast mobile
  robots in cluttered environments,'' in \emph{International Conference on
  Robotics and Automation}, 1990, pp. 572--577.

\bibitem{fox1997dynamic}
D.~Fox, W.~Burgard, and S.~Thrun, ``The dynamic window approach to collision
  avoidance,'' \emph{IEEE Robotics \& Automation Magazine}, vol.~4, no.~1, pp.
  23--33, 1997.

\bibitem{trautman2010unfreezing}
P.~Trautman and A.~Krause, ``Unfreezing the robot: Navigation in dense,
  interacting crowds,'' in \emph{IEEE/RSJ International Conference on
  Intelligent Robots and Systems}, 2010, pp. 797--803.

\bibitem{chen2017decentralized}
Y.~F. Chen, M.~Liu, M.~Everett, and J.~P. How, ``Decentralized
  non-communicating multiagent collision avoidance with deep reinforcement
  learning,'' in \emph{IEEE International Conference on Robotics and
  Automation}, 2017, pp. 285--292.

\bibitem{trautman2015robot}
P.~Trautman, J.~Ma, R.~M. Murray, and A.~Krause, ``Robot navigation in dense
  human crowds: Statistical models and experimental studies of human--robot
  cooperation,'' \emph{International Journal of Robotics Research}, vol.~34,
  no.~3, pp. 335--356, 2015.

\bibitem{matsumaru2006mobile}
T.~Matsumaru, T.~Kusada, and K.~Iwase, ``Mobile robot with
  preliminary-announcement function of forthcoming motion using light-ray,'' in
  \emph{IEEE/RSJ International Conference on Intelligent Robots and Systems},
  2006, pp. 1516--1523.

\bibitem{matsumaru2007mobile}
T.~Matsumaru, ``Mobile robot with preliminary-announcement and indication
  function of forthcoming operation using flat-panel display,'' in \emph{IEEE
  International Conference on Robotics and Automation}, 2007, pp. 1774--1781.

\bibitem{watanabe2015communicating}
A.~Watanabe, T.~Ikeda, Y.~Morales, K.~Shinozawa, T.~Miyashita, and N.~Hagita,
  ``Communicating robotic navigational intentions,'' in \emph{IEEE/RSJ
  International Conference on Intelligent Robots and Systems}, 2015, pp.
  5763--5769.

\bibitem{leibo2017multi}
J.~Z. Leibo, V.~Zambaldi, M.~Lanctot, J.~Marecki, and T.~Graepel, ``Multi-agent
  reinforcement learning in sequential social dilemmas,'' in
  \emph{International Conference on Autonomous Agents and Multi-Agent Systems},
  2017, pp. 464--473.

\bibitem{chen2019crowd}
C.~Chen, Y.~Liu, S.~Kreiss, and A.~Alahi, ``Crowd-robot interaction:
  Crowd-aware robot navigation with attention-based deep reinforcement
  learning,'' in \emph{IEEE International Conference on Robotics and
  Automation}, 2019, pp. 6015--6022.

\bibitem{van2008reciprocal}
J.~Van~den Berg, M.~Lin, and D.~Manocha, ``Reciprocal velocity obstacles for
  real-time multi-agent navigation,'' in \emph{IEEE International Conference on
  Robotics and Automation}, 2008, pp. 1928--1935.

\bibitem{van2011reciprocal}
J.~Van Den~Berg, S.~J. Guy, M.~Lin, and D.~Manocha, ``Reciprocal n-body
  collision avoidance,'' in \emph{Robotics Research}.\hskip 1em plus 0.5em
  minus 0.4em\relax Springer, 2011, pp. 3--19.

\bibitem{ferrer2013robot}
G.~Ferrer, A.~Garrell, and A.~Sanfeliu, ``Robot companion: A social-force based
  approach with human awareness-navigation in crowded environments,'' in
  \emph{IEEE/RSJ International Conference on Intelligent Robots and Systems},
  2013, pp. 1688--1694.

\bibitem{tai2018socially}
L.~Tai, J.~Zhang, M.~Liu, and W.~Burgard, ``Socially compliant navigation
  through raw depth inputs with generative adversarial imitation learning,'' in
  \emph{IEEE International Conference on Robotics and Automation}, 2018, pp.
  1111--1117.

\bibitem{long2017deep}
P.~Long, W.~Liu, and J.~Pan, ``Deep-learned collision avoidance policy for
  distributed multiagent navigation,'' \emph{IEEE Robotics and Automation
  Letters}, vol.~2, no.~2, pp. 656--663, 2017.

\bibitem{liu2018map}
Y.~Liu, A.~Xu, and Z.~Chen, ``Map-based deep imitation learning for obstacle
  avoidance,'' in \emph{IEEE/RSJ International Conference on Intelligent Robots
  and Systems}, 2018, pp. 8644--8649.

\bibitem{pfeiffer2016predicting}
M.~Pfeiffer, U.~Schwesinger, H.~Sommer, E.~Galceran, and R.~Siegwart,
  ``Predicting actions to act predictably: Cooperative partial motion planning
  with maximum entropy models,'' in \emph{IEEE/RSJ International Conference on
  Intelligent Robots and Systems}, 2016, pp. 2096--2101.

\bibitem{chen2017socially}
Y.~F. Chen, M.~Everett, M.~Liu, and J.~P. How, ``Socially aware motion planning
  with deep reinforcement learning,'' in \emph{IEEE/RSJ International
  Conference on Intelligent Robots and Systems}, 2017, pp. 1343--1350.

\bibitem{everett2018motion}
M.~Everett, Y.~F. Chen, and J.~P. How, ``Motion planning among dynamic,
  decision-making agents with deep reinforcement learning,'' in \emph{IEEE/RSJ
  International Conference on Intelligent Robots and Systems}, 2018, pp.
  3052--3059.

\bibitem{fan2019getting}
T.~Fan, X.~Cheng, J.~Pan, P.~Long, W.~Liu, R.~Yang, and D.~Manocha, ``Getting
  robots unfrozen and unlost in dense pedestrian crowds,'' \emph{IEEE Robotics
  and Automation Letters}, vol.~4, no.~2, pp. 1178--1185, 2019.

\bibitem{chen2019robot}
Y.~Chen, C.~Liu, M.~Liu, and B.~E. Shi, ``Robot navigation in crowds by graph
  convolutional networks with attention learned from human gaze,'' \emph{arXiv
  preprint arXiv:1909.10400}, 2019.

\bibitem{robicquet2016learning}
A.~Robicquet, A.~Sadeghian, A.~Alahi, and S.~Savarese, ``Learning social
  etiquette: Human trajectory understanding in crowded scenes,'' in
  \emph{European Conference on Computer Vision}, 2016, pp. 549--565.

\bibitem{claus1998dynamics}
C.~Claus and C.~Boutilier, ``The dynamics of reinforcement learning in
  cooperative multiagent systems,'' in \emph{AAAI Conference on Artificial
  Intelligence}, vol. 1998, no. 746-752, 1998, p.~2.

\bibitem{tan1993multi}
M.~Tan, ``Multi-agent reinforcement learning: Independent vs. cooperative
  agents,'' in \emph{International Conference on Machine Learning}, 1993, pp.
  330--337.

\bibitem{fisac2019hierarchical}
J.~F. Fisac, E.~Bronstein, E.~Stefansson, D.~Sadigh, S.~S. Sastry, and A.~D.
  Dragan, ``Hierarchical game-theoretic planning for autonomous vehicles,'' in
  \emph{IEEE International Conference on Robotics and Automation}, 2019, pp.
  9590--9596.

\bibitem{ma2017forecasting}
W.-C. Ma, D.-A. Huang, N.~Lee, and K.~M. Kitani, ``Forecasting interactive
  dynamics of pedestrians with fictitious play,'' in \emph{IEEE Conference on
  Computer Vision and Pattern Recognition}, 2017, pp. 774--782.

\bibitem{jaques2019social}
N.~Jaques, A.~Lazaridou, E.~Hughes, C.~Gulcehre, P.~Ortega, D.~Strouse, J.~Z.
  Leibo, and N.~De~Freitas, ``Social influence as intrinsic motivation for
  multi-agent deep reinforcement learning,'' in \emph{International Conference
  on Machine Learning}, 2019, pp. 3040--3049.

\bibitem{lanctot2017unified}
M.~Lanctot, V.~Zambaldi, A.~Gruslys, A.~Lazaridou, K.~Tuyls, J.~P{\'e}rolat,
  D.~Silver, and T.~Graepel, ``A unified game-theoretic approach to multiagent
  reinforcement learning,'' in \emph{Advances in Neural Information Processing
  Systems}, 2017, pp. 4190--4203.

\bibitem{hardin1968tragedy}
G.~Hardin, ``The tragedy of the commons,'' \emph{Science}, vol. 162, no. 3859,
  pp. 1243--1248, 1968.

\bibitem{fraichard2020crowd}
T.~Fraichard and V.~Levesy, ``From crowd simulation to robot navigation in
  crowds,'' \emph{IEEE Robotics and Automation Letters}, vol.~5, no.~2, pp.
  729--735, 2020.

\bibitem{xu2019crowd}
M.~Xu, X.~Xie, P.~Lv, J.~Niu, H.~Wang, C.~Li, R.~Zhu, Z.~Deng, and B.~Zhou,
  ``Crowd behavior simulation with emotional contagion in unexpected
  multihazard situations,'' \emph{IEEE Transactions on Systems, Man, and
  Cybernetics: Systems}, 2019.

\bibitem{mnih2015human}
V.~Mnih, K.~Kavukcuoglu, D.~Silver, A.~A. Rusu, J.~Veness, M.~G. Bellemare,
  A.~Graves, M.~Riedmiller, A.~K. Fidjeland, G.~Ostrovski \emph{et~al.},
  ``Human-level control through deep reinforcement learning,'' \emph{Nature},
  vol. 518, no. 7540, pp. 529--533, 2015.

\bibitem{openaigym}
G.~Brockman, V.~Cheung, L.~Pettersson, J.~Schneider, J.~Schulman, J.~Tang, and
  W.~Zaremba, ``Openai gym,'' 2016.

\bibitem{kingma2014adam}
D.~P. Kingma and J.~Ba, ``Adam: A method for stochastic optimization,''
  \emph{arXiv preprint arXiv:1412.6980}, 2014.

\bibitem{brambilla2013swarm}
M.~Brambilla, E.~Ferrante, M.~Birattari, and M.~Dorigo, ``Swarm robotics: a
  review from the swarm engineering perspective,'' \emph{Swarm Intelligence},
  vol.~7, no.~1, pp. 1--41, 2013.

\bibitem{cao2010distributed}
Y.~Cao, D.~Stuart, W.~Ren, and Z.~Meng, ``Distributed containment control for
  multiple autonomous vehicles with double-integrator dynamics: algorithms and
  experiments,'' \emph{IEEE Transactions on Control Systems Technology},
  vol.~19, no.~4, pp. 929--938, 2010.

\end{thebibliography}

\end{document}